%% file: bmvc_review.tex
\documentclass{bmvc2k}

\title{Mask-aware IoU for Anchor Assignment in Real-time Instance Segmentation}

\addauthor{Kemal Oksuz$^\dagger$, Baris Can Cam$^\dagger$, \\ Fehmi Kahraman, Zeynep Sonat Baltaci, \\ Sinan Kalkan$^\ddagger$, Emre Akbas$^\ddagger$}{{kemal.oksuz,can.cam,fehmi.kahraman_01,sonat.baltaci,skalkan, eakbas}@metu.edu.tr}{1}
\addinstitution{
 Dept. of Computer Engineering\\
 Middle East Technical University\\
 Ankara, Turkey
}

\usepackage{amsthm}
\usepackage{floatrow}
\usepackage{wrapfig,lipsum,booktabs}

\newtheorem{definition}{Definition}
\usepackage{algpseudocode}
\usepackage{algorithm}
\usepackage{multirow}
\usepackage{caption,subcaption}
\usepackage{mwe}
\usepackage{cancel}
\usepackage{amsmath}
\usepackage{amsfonts}
\usepackage{comment}
\newcommand{\blockcomment}[1]{}

\usepackage{amssymb}% http://ctan.org/pkg/amssymb
\usepackage{pifont}% http://ctan.org/pkg/pifont
\newcommand{\cmark}{\ding{51}}%
\newcommand{\xmark}{\ding{55}}%
\usepackage{float}
\usepackage{url}
\floatstyle{plaintop}
\restylefloat{table}

\definecolor{bcolor}{RGB}{50, 125, 50}
\definecolor{forestgreen}{rgb}{0.13, 0.55, 0.13}

\runninghead{Oksuz et al.}{Mask-aware Intersection-over-Union}

\begin{document}

\maketitle

\begin{abstract}
This paper presents Mask-aware Intersection-over-Union (maIoU) for assigning anchor boxes as positives and negatives during training of instance segmentation methods. Unlike conventional IoU or its variants, which only considers the proximity of two boxes; maIoU consistently measures the proximity of an anchor box with not only a ground truth box but also its associated ground truth mask. Thus, additionally considering  the mask, which, in fact, represents the shape of the object, maIoU enables a more accurate supervision during training. We present the effectiveness of maIoU on a state-of-the-art (SOTA) assigner, ATSS, by replacing IoU operation by our maIoU and training YOLACT, a SOTA real-time instance segmentation method. Using ATSS with maIoU consistently outperforms (i) ATSS with IoU by $\sim 1$ mask AP, (ii) baseline YOLACT with fixed IoU threshold assigner by $\sim 2$ mask AP over different image sizes and (iii) decreases the inference time by $25 \%$ owing to using less anchors. Then, exploiting this efficiency, we devise maYOLACT, a faster and $+6$ AP more accurate detector than YOLACT. Our best model achieves $37.7$ mask AP at $25$ fps on COCO test-dev establishing a new state-of-the-art for real-time instance segmentation. Code is available at \url{https://github.com/kemaloksuz/Mask-aware-IoU}.
\end{abstract}

%-------------------------------------------------------------------------

\input{sections/1.Introduction}
\input{sections/2.RelatedWork}
\input{sections/3.mIoUAssigner}
\input{sections/4.Experiments}
\input{sections/5.Conclusions}

\noindent \textbf{Acknowledgements:} This work was supported by the Scientific and Technological Research Council of Turkey (T\"UB\.{I}TAK)  (under grants 117E054 and 120E494). We also gratefully acknowledge the computational resources kindly provided by T\"UB\.{I}TAK ULAKBIM High Performance and Grid Computing Center (TRUBA) and Roketsan Missiles Inc. Dr. Kalkan is supported by the BAGEP Award of the Science Academy, Turkey.

\bibliography{egbib}
\end{document}

%% file: sections/1.Introduction.tex
\section{Introduction}
\label{sec:Introduction}
\begin{figure}
    \centerline{
        \includegraphics[width=\textwidth]{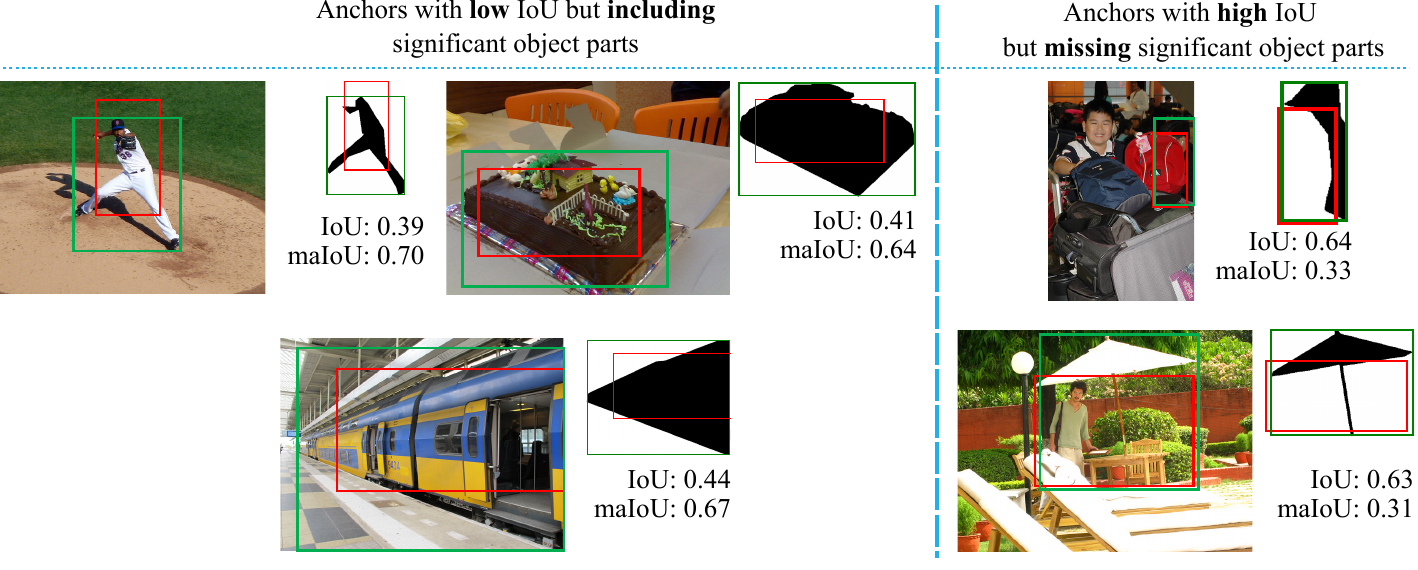}
    }
    \caption{Sample cases illustrating the need for mask-aware IoU (maIoU). Green boxes denote ground truth, red boxes are real anchors produced during the training. Left panel shows cases where the anchor covers a significant part of the object pixels but IoU is low (i.e. less than positive threshold of 0.50 for YOLACT). maIoU is higher than IoU for these cases, potentially correcting the assignment. Right panel shows cases where the anchor covers only a small part of the object pixels but IoU is high (so, anchors are positive). maIoU is lower than IoU, potentially correcting the assignment. Images are from COCO \cite{COCO}.\label{fig:Teaser}
} 
\end{figure}%
% The Assignment Problem in Instance Segmentation 
Instance segmentation is a visual detection problem which aims to classify and locate each object in an image by pixel-level masks. To be able to handle objects of different numbers, locations and scales; SOTA methods \cite{yolact,MaskRCNN,solov2} employ a dense set of object hypotheses, generally represented by boxes or points, and ensure a maximum coverage of the objects. This coverage necessitates a large number of object hypotheses ($\sim 20k$ per image in YOLACT \cite{yolact} for images of size $550 \times 550$) to be assigned to  ground truths boxes;  generally known as the \textit{assignment problem} \cite{ATSS,paa}.
%However, this coverage necessitates a large number of object hypotheses ($\sim 20k$ per image in YOLACT \cite{yolact} for images of size $550 \times 550$), which makes the manual assignment of hypotheses with the ground truths infeasible; generally known as the \textit{assignment problem} \cite{ATSS,paa}.

% A common heuristic 
This assignment problem is commonly tackled by employing heuristic rules. One common rule to assign object hypotheses represented by boxes (i.e. \textit{anchors}) is using a ``fixed IoU threshold''  \cite{yolact,MaskRCNN,retinamask}, in which an anchor, $\hat{B}$, can be assigned with a ground truth (i.e positive), $B$, when their Intersection-over-Union (IoU), defined as $\mathrm{IoU}(\hat{B}, B)=|\hat{B} \cap B|/|\hat{B} \cup B|$, exceeds a pre-determined threshold, $\tau$. If anchor $\hat{B}$ cannot be assigned to any ground-truth $B$ (i.e.  $\mathrm{IoU}(\hat{B}, B) < \tau,\;\,\forall B$), then $\hat{B}$ is assumed to be a background (i.e. negative) example. A different set of recent methods showed for object detection \cite{ATSS,paa,noisyanchor} that assigning the anchors using an ``adaptive IoU threshold'' determined for each ground truth improves the performance. Still, assignment methods heavily rely on IoU as the de facto proximity measure between ground truths and anchors.

% The problem with this strategy
Despite its popularity, IoU has a certain drawback: The IoU between an anchor box and a ground truth box solely depends on their areas, thereby ignoring the shape of the object, e.g. as  provided by a segmentation mask. This may give rise to undesirable assignments due to counter-intuitively lower or higher IoU scores. For example, the IoU might be high, implying a positive anchor, but only a small part of the object is included in the anchor; or the IoU may be low, implying a negative anchor, but a large part of the object is included in the anchor. Fig. \ref{fig:Teaser} presents examples for such cases, arising due to objects with unconventional poses, occlusion and objects with articulated or thin parts. We will show (in Section \ref{subsec:Analysis}, Fig. \ref{fig:MOB}) that such examples tend to produce larger loss values and adversely affect training.

% How do we address this?
In this paper, we introduce mask-aware IoU (maIoU), a novel IoU measure for anchor assignment in instance segmentation by exploiting the ground truth masks of the objects, normally used only for supervision. Specifically, unlike IoU, which equally weights all pixels, maIoU yields a proximity measure between $0$ and $1$ among an anchor box, a ground truth box and a ground truth mask by promoting the pixels on the masks, thereby providing a more consistent assignment (Fig. \ref{fig:Teaser}). Since a naive computation of maIoU is impractical, we present an efficient algorithm with training time similar to the baseline. YOLACT with maIoU-based ATSS assigner  consistently improves ATSS assigner with IoU by $\sim 1$ mask AP and standard YOLACT (i.e. fixed IoU threshold) by $\sim 2$ mask AP, and also decreases inference time of YOLACT. Finally, utilizing this efficiency gap, we build maYOLACT detector reaching $37.7$ mask AP at $25$ fps and outperforming all real-time counterparts.

%% file: sections/2.RelatedWork.tex
\section{Related Work}
\label{sec:RelatedWork}
\begin{table}
    \centering
    \small
    \caption{\small IoU Variants, their inputs and primary purposes. IoU variants assign a proximity measure based on the properties (prop.) of two inputs (Input 1 and Input 2). In practice, existing variants compare the inputs wrt. the same properties (i.e. either boxes or masks). Our Mask-aware IoU (maIoU) can uniquely compare a box with a box and a mask. With this, maIoU compares anchors (i.e. only box) with ground truths (box and mask) in order to provide better anchor assignment. *: GIoU is also used as a performance measure. }
    \begin{tabular}{|c|c|c|c|c|c|}
         \hline
         \multirow{2}{*}{IoU Variant}&
         \multicolumn{2}{c|}{Input 1 prop.} & \multicolumn{2}{c|}{Input 2 prop.} & Primary purpose as \\
         \cline{2-5}
         &Box&Mask&Box&Mask&proposed in the paper\\ \hline \hline
         Mask IoU \cite{COCO, Cityscapes}& \xmark & \cmark & \xmark & \cmark  & Performance measure  \\ \hline 
         Boundary IoU \cite{boundaryiou}& \xmark & \cmark & \xmark & \cmark & Performance measure \\ \hline
         Generalized IoU \cite{GIoULoss}  & \cmark &\xmark & \cmark &  \xmark & Loss function* \\ \hline
         Distance IoU \cite{DIoULoss} & \cmark &\xmark & \cmark &  \xmark & Loss function \\ \hline
         Complete IoU \cite{CIoULoss} &\cmark &\xmark & \cmark &  \xmark & Loss function \\ \hline \hline
         \textit{Mask-aware IoU (Ours)} & \cmark & \cmark & \cmark & \xmark & \textit{Assigner} \\ \hline
    \end{tabular}
    \label{tab:iou_variants}
\end{table}
\noindent \textbf{Deep Instance Segmentation.} In general, deep instance segmentation methods have followed object detection literature. The pioneering Mask R-CNN model \cite{MaskRCNN} and its variations \cite{maskscoring,RSLoss} extended Faster R-CNN \cite{FasterRCNN} by incorporating a mask prediction branch into the two-stage detection pipeline. Similarly,  anchor-based one-stage methods were also adapted for instance segmentation by using an additional mask prediction branch, e.g. YOLACT \cite{yolact} and YOLACT++ \cite{yolact-plus} employed a YOLO-like architecture; and PolarMask \cite{polarmask} and PolarMask++ \cite{PolarMask-plus}, both anchor-free methods, adapted FCOS \cite{FCOS} for instance segmentation. Differently, SOLO variants \cite{solo,solov2} classify the pixels based on  location and size of each instance.

\noindent \textbf{Anchor Assignment in Instance Segmentation.} Deep instance segmentation methods using anchors as object hypotheses label anchors based on ``fixed IoU threshold'' assignment rule: The anchors with IoU larger than $\tau^+$ with a ground truth box are assigned as positive; while the anchors whose maximum IoU with ground truths is less than $\tau^-$ are assigned as negatives; and the remaining anchors whose maximum IoU is between $\tau^-$ and $\tau^+$ are simply ignored during training. To illustrate, YOLACT variants \cite{yolact,yolact-plus} and RetinaMask \cite{retinamask} use  $\tau^-=0.40$ and $\tau^+=0.50$; while the first stage of Mask R-CNN (i.e. region proposal network \cite{FasterRCNN}) sets $\tau^-=0.30$ and $\tau^+=0.70$; and finally its second stage \cite{MaskRCNN} uses $\tau^-=\tau^+=0.50$. 

%Recently, Zhang et al. \cite{ATSS} showed that representing these hypotheses either by anchors (i.e. bounding boxes) or by points (i.e. points) does not effect the performance and how these hypotheses are labelled as positives and negatives makes the difference. In this paper, we are interested in labelling the hypotheses represented as anchors.  

\noindent \textbf{Adaptive Anchor Assignment Methods in Object Detection.} Recently, adaptive anchor assignment strategies are shown to perform better than fixed IoU threshold in object detection: ATSS \cite{ATSS} uses top-k anchors wrt. IoU to determine an adaptive IoU threshold for each ground truth (Section \ref{subsec:ATSS} provides more details on ATSS.) and PAA \cite{paa} computes a score of each anchor including Generalized IoU and fits the distribution of these scores to a two-dimensional Gaussian Mixture Model to split positives and negatives for each ground truth. Similarly, Ke et al. \cite{mal} and Li et al. \cite{noisyanchor} identify  positives and negatives by using different scoring functions of the predictions. However, these methods are devised and tested for object detection, and thus, do not utilize object masks. 

\noindent \textbf{Other IoU Variants.} Over the years, many IoU variants have been proposed -- see Table \ref{tab:iou_variants} for a comparative summary. One of the most related IoU variants is Mask IoU, which is used to measure the detection mask quality with respect to (wrt.) the ground truth mask during evaluation \cite{COCO,Cityscapes}. Similarly, Boundary IoU \cite{boundaryiou} evaluates detection masks by giving higher weights to  the pixels closer to the boundaries. Note that these IoU variants compare only two masks, and unlike our maIoU, they cannot compare a box  with another box and its associated mask. The other IoU variants are all devised to measure the proximity of two boxes: Generalized IoU (GIoU) \cite{GIoULoss} uses the minimum enclosing box in order to measure the proximity of boxes when boxes do not intersect (i.e. their IoU is $0$); Distance IoU \cite{DIoULoss} adds a penalty parameter based on the minimum enclosing box and the distance between the centers of boxes; Complete IoU \cite{CIoULoss} additionally considers aspect ratio differences of the boxes. These IoU-variants compute the overlap at the box level and neglect object shape; and also, they are mainly used as loss functions, not as a positive-negative assignment criterion.
\begin{comment}
\begin{table}[h!]
    \centering
    \begin{tabular}{|c|c|c|}
         \hline
         \textbf{Method} & \textbf{G.T.} & \textbf{Anchor} \\ \hline
         Mask IoU \cite{COCO, Cityscapes} & Mask & Mask \\ 
         Bounded IoU \cite{boundaryiou} & Mask & Mask \\ \hline
         GIoU \cite{GIoULoss} & Box & Box \\ 
         DIoU \cite{DIoULoss} & Box & Box \\ 
         CIoU \cite{DIoULoss} & Box & Box \\ \hline
         maIoU (This Work) & Box + Mask & Box \\ \hline
    \end{tabular}
    \caption{Summary of IoU variants considering their annotation types for both ground-truth (G.T.) and anchor boxes. The only method that uses both mask and box annotations is maIoU.}
    \label{tab:iou_variants}
\end{table}
\end{comment}

\noindent \textbf{Comparative Summary.} By measuring the proximity of an anchor box with a ground truth, our maIoU is designed for anchor-based models, which have been using a fixed IoU threshold for assigning anchors as the dominant approach, thereby ignoring  object shape. We first show that ATSS \cite{ATSS}, an adaptive anchor assigner, yields better performance on YOLACT \cite{yolact}. Then, we propose maIoU, as the first IoU variant that can measure the proximity of an anchor box with a ground truth box and ground truth mask (Table \ref{tab:iou_variants}). Replacing IoU of ATSS by our maIoU improves the performance of this strong baseline. We also investigate  GIoU and DIoU for anchor assignment. Since they rely only on boxes, our maIoU provides more discriminative information then these IoU variants. Finally, besides our maIoU, adopting recently proposed improvements into YOLACT detector, we build maYOLACT detector, which outperforms its counterparts while being more efficient as well (Section \ref{sec:Experiments}).

%% file: sections/3.mIoUAssigner.tex
\section{Methodology}
\label{sec:miou}

This section first presents an analysis on fixed-threshold IoU assigner in Section \ref{subsec:Analysis}. Then, Section \ref{subsec:maIoU} defines maIoU and Section \ref{subsec:Algorithm} provides an efficient algorithm to compute maIoU. Finally, Section \ref{subsec:ATSS} incorporates our maIoU into the SOTA ATSS assigner \cite{ATSS} to label  anchors during the training of instance segmentation methods.

\begin{figure}
    \centering
    \includegraphics[width=\textwidth]{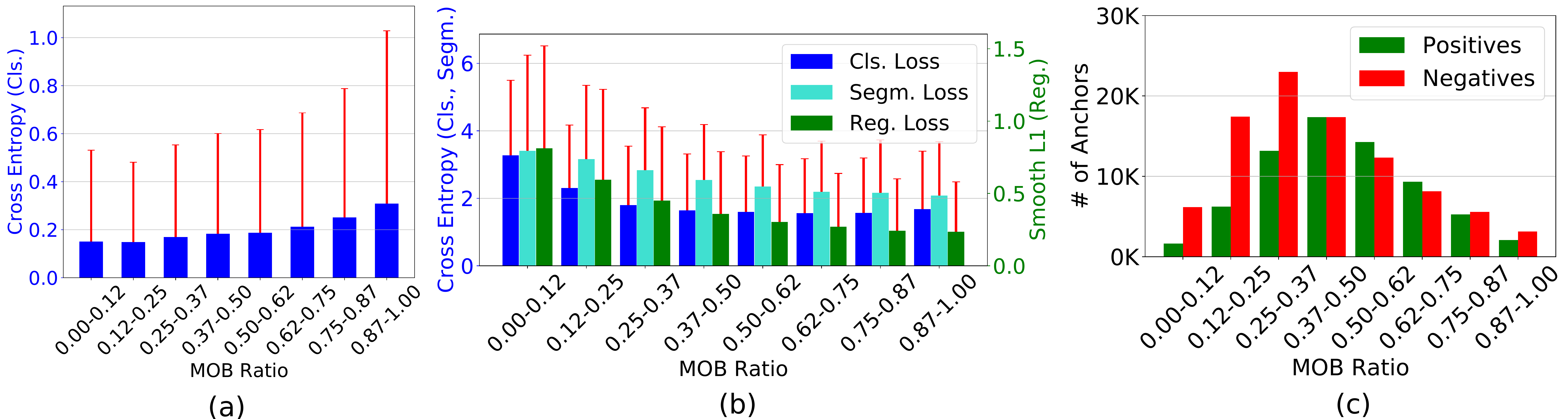}
    \caption{\small \textbf{(a,b)} Mean and standard deviation of loss values of negative anchors (a) and positive (b) anchors with \textbf{similar IoUs} (IoU between $[0.30-0.50]$ for negatives and $[0.50-0.70]$ for positives) over different MOB ratios for a \textbf{trained} YOLACT on COCO \textit{minival}. Red lines denote the standard deviation. Note that when the MOB ratio increases, the loss values  increase for negatives; however,  the loss values of all three sub-tasks (Cls.: classification, Segm.: segmentation, Reg.: regression) tend to decrease for positives.
    %The examples with larger losses are outliers, and tend to destabilize and dominate training.
    \textbf{(c)} The number of anchors for each MOB ratio is in the order of thousands.
    \label{fig:LossAnalysis}}
\end{figure}

\subsection{The Mask-over-box Ratio}
\label{subsec:Analysis}

Our analysis is based on an intuitive measure to represent the rate of the ground truth mask in a box, defined as the Mask-over-box (MOB) ratio as follows.

\begin{definition}
Mask-over-box (MOB) ratio of a box (i.e. anchor or ground truth), $\bar{B}$, on a ground truth mask, $M$, is the ratio of (i) the area of the intersection of the mask and the box, and (ii) the area of $\bar{B}$ itself: %
\begin{align}
\mathrm{MOB}(\bar{B},M)= \frac{|\bar{B} \cap M|}{|\bar{B}|}.     
\end{align}
\end{definition}

$\mathrm{MOB}(\bar{B},M) \in [0,1]$  and less object pixels from $M$ in $\bar{B}$ implies a lower MOB ratio. When $\bar{B}$ is the ground truth box of $M$ (i.e. $\bar{B}=B$), $|M \cap B|=|M|$, and thus $\mathrm{MOB}(B,M)=|M|/|B|$. 

We now analyse how losses on a trained standard YOLACT model (with ResNet-50 backbone), a SOTA instance segmentation method, and the number of anchors change wrt. MOB ratio (Fig. \ref{fig:LossAnalysis}) and make the following crucial observations:

\noindent \textbf{(1) The loss value (i.e. error) of a trained YOLACT for an anchor is related to the amount of mask pixels covered by that anchor (i.e. MOB ratio), which is ignored by the standard fixed IoU threshold assigner.} Fig. \ref{fig:LossAnalysis}(a,b) present that the average and standard deviation of the loss values of anchors that are close to the IoU assignment threshold (i.e. the anchors with IoU between $[0.30-0.50]$ for negatives and $[0.50-0.70]$ for positives; hence, anchors with similar IoUs) \textit{increase in all tasks} (for negatives, it is only classification; for positives, we look at all tasks -- i.e. classification, box regression and segmentation) as MOB ratio decreases/increases (i.e. implying covering less/more on-mask pixels) for positive/negative anchors. 
%We present the average and standard deviation of the loss values of anchors over their MOB ratios around the IoU assignment boundary . 
%While the anchors have similar IoUs, the trained model yields larger errors and standard deviations in all tasks when a positive/negative anchor has a smaller/larger MOB ratio, implying covering less/more on-mask pixels. 
Also, the numbers of anchors with larger losses are in the order of thousands in all cases (Fig. \ref{fig:LossAnalysis}(c)). 
%This suggests that MOB ratio has an effect on how accurate an anchor can be predicted by the trained model and IoU does not discriminate anchors with more on-mask pixels from those with less on-mask pixels.

\noindent \textbf{(2) Similar to anchors, MOB ratios of the ground truth boxes also vary significantly}. We observe in Fig. \ref{fig:MOB}(a) that there exists a significant amount of ground truth boxes with low MOB ratios (i.e. for 30 \% of the ground truths, MOB ratio is less than $0.50$. 

Our observations suggest that IoU does not discriminate anchors with more on-mask pixels from those with less on-mask pixels, which appears naturally due to varying MOB ratios of the ground truths; and thus using IoU for the assignment of the anchors may not be the best method. 

%IoU is not able to capture different degrees of overlap of anchors with ground truth boxes and masks, and Fig. \ref{fig:LossAnalysis} suggests this can have a significant effect on the quality of assignment.

%In order assure that our hypothesis generalizes over the dataset and not limited to some specific images,  of the ground truths in the training set and plot their distribution in Fig. \ref{fig:Definition}(b).  Overall, we observe the following: (i) Treating every box equally, as in IoU, may not be the best method for assignment since MOB ratios vary significantly. (ii) There exists a significant amount of boxes with low MOB ratios (i.e. for 30 \% of the ground truths, MOB ratio is less than $0.50$). These two observations confirm that our claims are not limited to few limited ground truths but many examples are affected during assignment.

\subsection{Mask-aware Intersection-over-Union (maIoU)}
\label{subsec:maIoU}
\begin{figure}
        \captionsetup[subfigure]{}
        \centering
        \begin{subfigure}[b]{0.35\textwidth}
                \includegraphics[width=\textwidth]{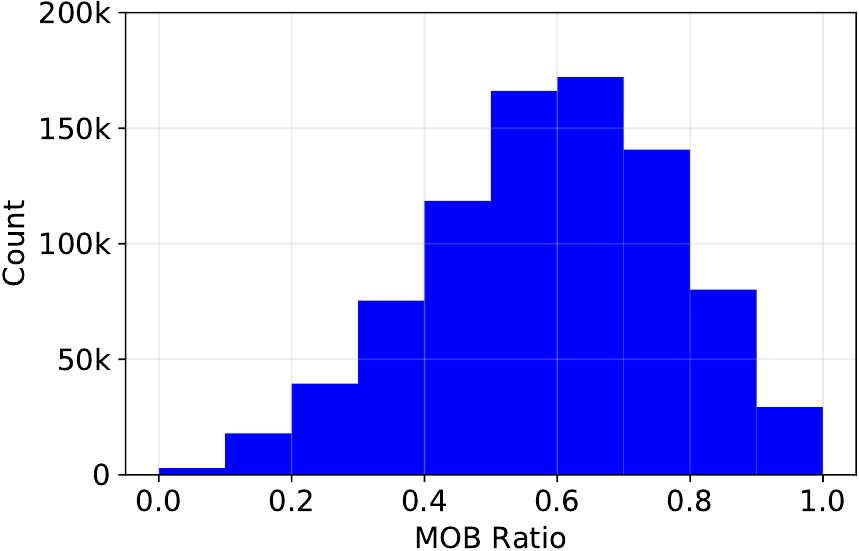}
                \caption{}
        \end{subfigure}       
        \begin{subfigure}[b]{0.3\textwidth}
                \includegraphics[width=\textwidth]{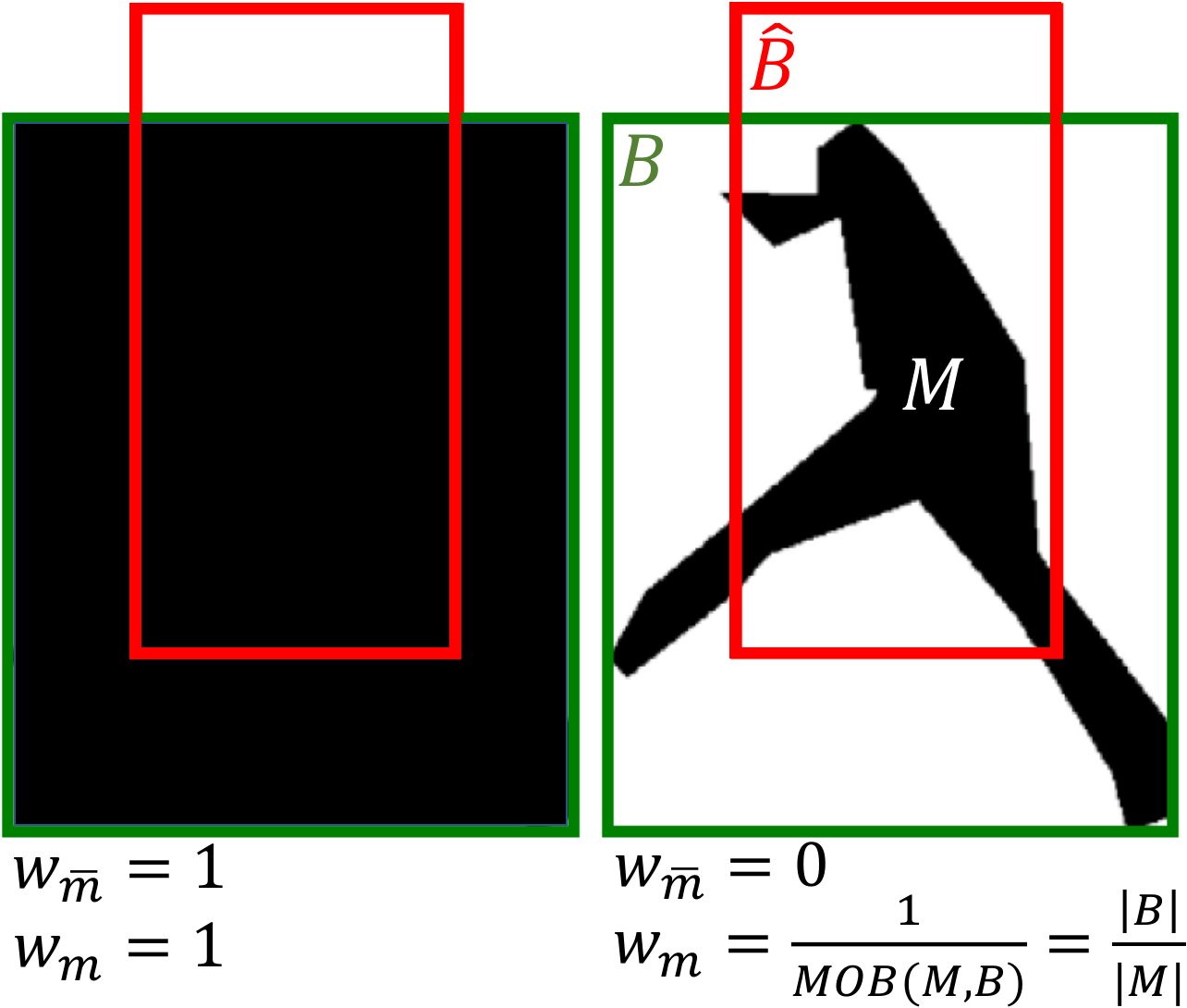}
                %\vspace{0mm}
                \caption{}
        \end{subfigure}
        \begin{subfigure}[b]{0.27\textwidth}
                \includegraphics[width=\textwidth]{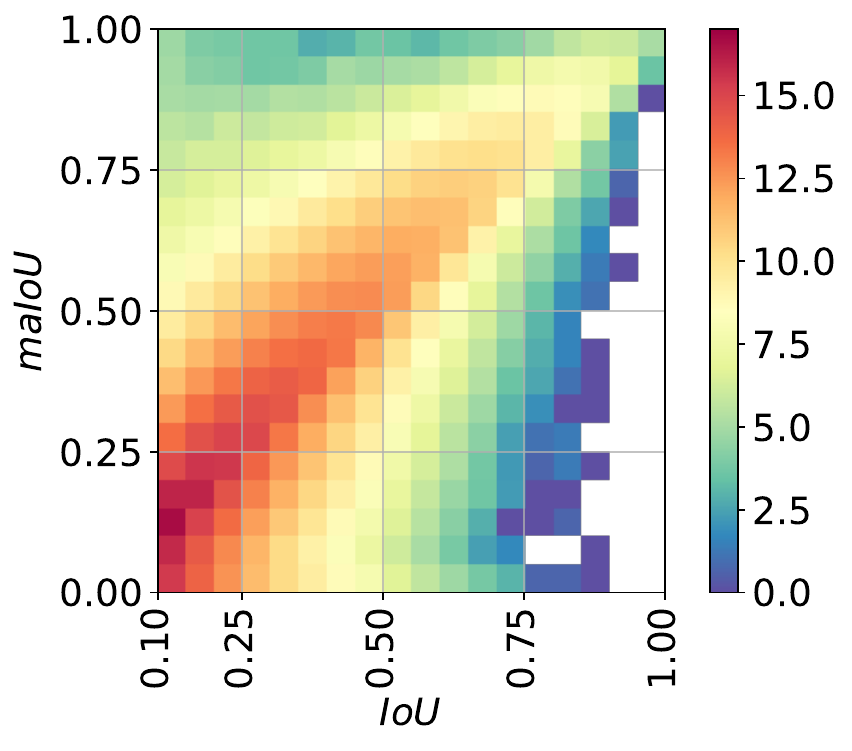}
                \caption{}
        \end{subfigure}             
        \caption{(a) The distribution of MOB ratios of the ground truths on COCO training set. (b) How IoU and maIoU weights the pixels in ground truth box (see Fig. \ref{fig:Teaser} top-left example for the image of this example). While IoU does not differentiate among on-mask ($w_m$) and off-mask ($w_{\overline{m}}$) pixels, our maIoU sets $w_{\overline{m}}=0$ and weights each on-mask pixel by inverse MOB ratio considering object mask $M$. (c) Anchor count distribution (in log-scale) of IoU vs maIoU. While $\mathrm{IoU}$ \& high-$\mathrm{maIoU}$ positively correlate, there are quite a few examples with low-$\mathrm{IoU}$ \& high-$\mathrm{maIoU}$ and vice versa.}
        \label{fig:MOB}
\end{figure}

\noindent \textbf{Intuition.} The main intuition behind mask-aware IoU (maIoU) is to reweigh the pixels within the ground truth box, $B$, such that on-mask pixels are promoted (in a way, the contribution of off-mask pixels are reduced) by preserving the total \textit{energy} of $B$ (i.e. $|B|$). We simply achieve this by distributing the contributions of the off-mask pixels uniformly over the on-mask pixels in $B$. Finally, maIoU is computed as an intersection-over-union between $\hat{B}$ (i.e. anchor box) and $B$ with the new pixel weights in $B$ (Fig. \ref{fig:MOB}(b)).

\noindent \textbf{Derivation.} To facilitate derivation, we first reformulate intersection $\mathcal{I}$ between $B$ and $\hat{B}$ in a weighted form as follows ($w_m, w_{\overline{m}}$: the contributions of on-mask and off-mask pixels respectively):
\begin{equation}
\mathcal{I}(B, \hat{B}) = \sum_{i\ \in\ B \cap \hat{B}} w = \sum_{i\ \in\ \hat{B} \cap M} w_m +  \sum_{i\ \in\ \left(\hat{B} \cap B - \hat{B} \cap M \right )} w_{\overline{m}}, \label{eqn:IoU}
\end{equation}
which effectively does not make use of the mask $M$ since $w=w_m=w_{\overline{m}}=1$ for IoU. 

In maIoU, we discard the contribution of an off-mask pixel: $w_{\overline{m}}=0$, and in order to preserve the total energy, the reduced contribution from all off-mask pixels, which equals $|B|-|M|$, is distributed to the on-mask pixels uniformly. This will increase $w_m$ by $(|B|-|M|)/|M|$: $w_m=1+(|B|-|M|)/|M|=1+|B|/|M|-|M|/|M|=|B|/|M|=1/\mathrm{MOB}(B,M)$. 

With these weights, the mask-aware intersection, $\mathrm{ma}\mathcal{I}$ is defined by extending Eq. \ref{eqn:IoU}:
\begin{align}
     \mathrm{ma}\mathcal{I} (\hat{B}, B, M) & = \sum_{i\ \in\ \hat{B} \cap M} w_m +  \cancelto{0}{\sum_{i\ \in\ \left( \hat{B} \cap B-\hat{B} \cap M \right )} w_{\overline{m}}}= w_m |\hat{B} \cap M| = \frac{ 1}{\mathrm{MOB}(B,M)}|\hat{B} \cap M|.
\end{align}
Extending the definition of union (i.e. $|B|+|\hat{B}|-|B \cap \hat{B}|$) with this intersection concept:
\begin{align}
    \mathrm{ma}\mathcal{U} (\hat{B}, B, M) = |B| + (|\hat{B}|-|\hat{B} \cap B| + \mathrm{ma}\mathcal{I}(\hat{B}, B, M)) - \mathrm{ma}\mathcal{I} (\hat{B}, B, M) = |\hat{B} \cup B|,
\end{align}
which is equal to the conventional union. This is not surprising since our formulation preserves ground truth area (i.e. $|B|$). With the updated definitions, mask-aware IoU is simply mask-aware intersection over mask-aware union:
\begin{align}
    \label{eq:Definition1}
    \mathrm{maIoU} (\hat{B}, B, M) = \frac{\mathrm{ma}\mathcal{I} (\hat{B}, B, M)}{\mathrm{ma}\mathcal{U} (\hat{B}, B, M)} =  \frac{1}{\mathrm{MOB}(B,M)}\frac{|\hat{B} \cap M|}{|\hat{B} \cup B|},
\end{align}
which, in effect, is the ratio of covered on-mask pixels by the anchor ($|\hat{B} \cap M|$) in the union of boxes ($|\hat{B} \cup B|$), normalized by on-mask pixel density in $B$ (i.e. $\mathrm{MOB}(B,M)$).

\noindent \textbf{Interpretation.} Similar to IoU, $\mathrm{maIoU} (\hat{B}, B, M) \in [0,1]$ and a larger value implies a better localisation considering not only the boxes but also the ground truth mask (Fig. \ref{fig:MOB}(b)). We visualize the anchor count  distribution (in log-scale) of YOLACT on COCO minival on the space spanned by $\mathrm{IoU}$ and $\mathrm{maIoU}$ (Fig. \ref{fig:MOB}(c)): While $\mathrm{IoU}$ and $\mathrm{maIoU}$ are positively correlated, there are quite a number of examples with low-$\mathrm{IoU}$ \& high-$\mathrm{maIoU}$ and vice versa, hence the assignment rules based on $\mathrm{maIoU}$ is quite different than those based on $\mathrm{IoU}$.  
%
%\begin{figure}
%    \centering
%    \includegraphics[width=0.99\textwidth]{figures/space_of_mIoU.pdf}
%    \caption{A comparison of IoU and maIoU on how the anchors are distributed and split as positive, negative and ignored. \label{fig:the_space_of_mIoU} }
%\end{figure}

%\begin{figure*}[t]
%        \captionsetup[subfigure]{}
%        \centering
%        \begin{subfigure}[b]{0.24\textwidth}
%        \includegraphics[width=\textwidth]{figures/loss_analysis/fig_a.pdf}       
%        \caption{Negative Examples Classification Loss}
%        \end{subfigure}
%        \begin{subfigure}[b]{0.54\textwidth}
%        \includegraphics[width=\textwidth]{figures/loss_analysis/fig_b.pdf}       
%        \caption{Positive Examples Losses}
%        \end{subfigure}
%        \begin{subfigure}[b]{0.24\textwidth}
%        \includegraphics[width=\textwidth]{figures/loss_analysis/fig_c.pdf}
%        \caption{Anchor Counts}
%        \end{subfigure}
%        \caption{}
%        \label{fig:LossAnalysis}
%\end{figure*}

\subsection{Computation of maIoU}
\label{subsec:Algorithm}

Compared to IoU, computing maIoU involves two additional terms (Eq. \ref{eq:Definition1}):  (i) $|M|$, the total number of mask-pixels (since calculating $|B|$ for $\mathrm{MOB}(B,M)$=$|M|/|B|$ is trivial), and (ii) $|\hat{B} \cap M|$, the number of mask-pixels in the intersection. While the masks are included in the datasets and computing these two quantities is straightforward; it is impractical to compute them naively (i.e. brute force, see Table \ref{tab:runtime}) considering the large number of anchors covering the image. For this reason, we employ \textit{integral images} \cite{IntegralImage} on the binary ground truth masks, $M$. More specifically, for $M$ covering an image of size $m \times n$, we compute its integral image $\Theta^M$, an $(m+1) \times (n+1)$ matrix that encodes the total number of mask pixels above and to the left of each pixel. Accordingly, denoting the $(i,j)^{th}$ element of $\Theta^M$ by $\Theta^M_{i,j}$, the last element of $\Theta^M$ stores $|M|$, i.e. $|M|=\Theta^M_{m+1,n+1}$. As for the second term, assuming $\hat{B}$ is represented by a top-left point $(x_1, y_1)$ and a bottom-right point $(x_2,y_2)$ such that $x_2>x_1$ and $y_2>y_1$, $|\hat{B} \cap M|$ involves only four look-up operations, i.e. $|\hat{B} \cap M|=\Theta^M_{x_2+1,y_2+1}+\Theta^M_{x_1,y_1}-\Theta^M_{x_2+1,y_1}-\Theta^M_{x_1,y_2+1}$. The overall algorithm to compute $\mathrm{maIoU}(\hat{B}, B, M)$ is presented in Alg. \ref{alg:MaskAwareIoU}.
\begin{algorithm}
\caption{The algorithm for efficiently calculating mask-Aware IoU. \label{alg:MaskAwareIoU}}
\small
\begin{algorithmic}[1]
\Procedure{MaskAwareIoU}{$\hat{B}, B, M$}
\State Compute $|B|$, $|\hat{B} \cup B|$ and $\Theta^M$ as integral image of $M$ such that $\Theta^M_{i,j}$ is $(i,j)^{th}$ element of $\Theta^M$ 
\State Set $|M|=\Theta^M_{m+1,n+1}$ and $|\hat{B} \cap M|=\Theta^M_{x_2+1,y_2+1}+\Theta^M_{x_1,y_1}-\Theta^M_{x_2+1,y_1}-\Theta^M_{x_1,y_2+1}$.
\State Compute $\mathrm{MOB}(B,M)= |M|/|B|$ for ground truth $B$
\State \textbf{return} $\mathrm{maIoU} (\hat{B}, B, M)$ (Eq. \ref{eq:Definition1})
\EndProcedure
\end{algorithmic}
\end{algorithm}

% In order to improve the efficiency, we dynamically select for which anchors Mask-aware IoU computation is required by employing two useful points: Firstly, $\mathrm{IoU}(B, \hat{B})=0$ implies $\mathrm{mIoU}_\gamma (B, M, \hat{B})=0$. Secondly, in cases where a minimum matching threshold, $mIoU_{\gamma}^{min}$, is specified for positives \cite{FasterRCNN}, this IoU threshold can be set more tight than $0$. In particular, combining these two observations, we only compute $\mathrm{mIoU}$ for anchors with $IoU> \max\left(0, \frac{mIoU_{\gamma}^{min}-\gamma}{1-\gamma}\right)$ (See Supp.Mat for the derivation and Section \ref{subsec:efficiency} for the analysis). 

% The algorithm to compute $\mathrm{mIoU}$ between a single anchor and ground truth is presented in Algorithm \ref{alg:MaskAwareIoU}. Considering a detection pipeline involving many anchors and few ground truths in every training batch, the additional computation burden originates only from lines 5-9 since line 2 computes IoU. Lines 5-7 are executed for the number of ground truths times and do not have a significant impact. In line 8, for the anchors with IoU larger than the threshold, four look-up operations are executed in constant time thanks to integral images and subsequently line 9 computes Mask-aware IoU. Note that there is no additional burden at test time since positive-negative supervision is not possible.

\subsection{Incorporating maIoU into ATSS Assigner}
\label{subsec:ATSS}
ATSS assigner \cite{ATSS} is a SOTA assignment method used for object detection, yielding better performance than a fixed-threshold IoU assigner and simplifying the anchor design by using a single anchor per pixel unlike its predecessors with up to nine anchors per pixel \cite{FocalLoss}. ATSS assigner comprises three steps: (i) selecting top-$k$ (i.e. conventionally $k=9$) anchors ($\hat{B}$) wrt. the distance of the centers between $B$ and $\hat{B}$ for each ground truth ($B$) on each FPN level as ``candidates'', (ii) filtering out the candidates using an adaptive IoU threshold, computed based on the statistics of these candidates for each $B$, and (iii) filtering out the candidates, whose centers lie out of $B$. The surviving candidates after steps (ii) and (iii) are the positive examples and the remaining examples are the negatives. Using our maIoU with ATSS (or any IoU-based assigner) is straightforward: In step (ii), we just replace IoU-based adaptive thresholding by maIoU-based adaptive thresholding.

%% file: sections/4.Experiments.tex
\section{Experiments}
\label{sec:Experiments}

\noindent \textbf{Dataset.} We train all models on the COCO \textit{trainval} set \cite{COCO} (115K images), test them on the COCO \textit{minival} set (5k images) unless otherwise stated.

\noindent \textbf{Performance Measures.} We mainly report AP-based performance metrics: COCO-style AP (AP, in short), APs where true positives are validated from IoUs of 0.50 and 0.75 ($\mathrm{AP_{50}}$ and $\mathrm{AP_{75}}$), and APs for small, medium and large objects ($\mathrm{AP_{S}}$, $\mathrm{AP_{M}}$ and $\mathrm{AP_{L}}$ respectively). Furthermore, we also exploit the recent optimal Localisation Recall Precision (oLRP) Error \cite{LRP, LRParXiv}. While  AP is a  higher-is-better measure, oLRP is a lower-is-better metric.

\noindent \textbf{Implementation Details.} We conduct our experiments on YOLACT \cite{yolact}, an anchor-based real-time instance segmentation method, using the mmdetection framework \cite{mmdetection}. Following Zhang et al. \cite{ATSS}, when we use ATSS assigner, we keep $k=9$ (Section \ref{subsec:ATSS}) and simplify the anchor configuration by placing a single anchor on each pixel with an aspect ratio of $1:1$ and a base scale of $4$ unless otherwise stated. Also, with ATSS (with IoU, DIoU \cite{DIoULoss}, GIoU \cite{GIoULoss} or our  maIoU), we keep classification and box regression loss weights as they are ($1.0$ and $1.5$ respectively), and increase mask prediction loss weight from $6.125$ to $8$. When we replace the assigner, note that it affects the examples in all branches (i.e. classification, box regression, semantic and instance mask predictions). We train all models with $32$ images distributed on $4$ GPUs ($8$ images/GPU). The remaining design choices of YOLACT \cite{yolact} are kept. We adopt ResNet-50 \cite{ResNet} as the backbone and resize the images during training and inference to $S \times S$ where $S$ can be either 400, 550 or 700 following Bolya et al. \cite{yolact}.

\begin{table}
    \centering
    \small
    \setlength{\tabcolsep}{0.35em}
    \caption{Comparison of different assigners and IoU variants on YOLACT. Considering the shapes of the objects, our ATSS w. our maIoU consistently outperforms its  counterparts.}
    \label{tab:minival}
    \begin{tabular}{|c|c|c|c|c|c|c|c|c|} \hline
        Scale & Assigner & $\mathrm{AP} \uparrow$ & $\mathrm{AP_{50}} \uparrow$ & $\mathrm{AP_{75}} \uparrow$ & $\mathrm{AP_{S}} \uparrow$ & $\mathrm{AP_{M}} \uparrow$ & $\mathrm{AP_{L}} \uparrow$ & $\mathrm{oLRP} \downarrow$ \\ \hline
    \multirow{5}{*}{400}&fixed IoU threshold&$24.8$&$42.4$&$25.0$&$\mathbf{7.3}$&$26.0$&$42.0$&$78.3$ \\
    &ATSS w. IoU&$25.3$&$43.5$&$25.5$&$6.8$&$27.3$&$43.8$&$77.7$ \\
    &ATSS w. DIoU&$25.4$&$43.6$&$25.2$&$7.2$&$27.1$&$43.4$&$77.7$ \\
    &ATSS w. GIoU&$25.1$&$42.7$&$25.3$&$7.0$&$26.8$&$41.8$&$78.0$ \\ \cline{2-9}
%    &ATSS w. maskIoU &$\mathbf{26.5}$&$\mathbf{44.4}$&$\mathbf{27.0}$&$7.0$&$\mathbf{27.9}$&$\mathbf{45.7}$&$\mathbf{?}$ \\
    &ATSS w. maIoU (Ours)&$\mathbf{26.1}$&$\mathbf{44.3}$&$\mathbf{26.3}$&$7.2$&$\mathbf{28.0}$&$\mathbf{44.3}$&$\mathbf{77.1}$\\ \hline \hline 
    \multirow{5}{*}{550} &fixed IoU threshold& $28.5$ & $47.9$ & $29.4$ & $11.7$ & $31.8$ & $43.0$ & $75.2$ \\
    & ATSS w. IoU&$29.3$& $49.2$ & $30.2$ & $11.1$ & $33.0$ & $44.5$ & $74.5$ \\
    &ATSS w. DIoU&$29.5$&$49.5$&$30.1$&$11.7$&$33.2$&$44.9$&$74.4$ \\
    &ATSS w. GIoU&$29.1$&$48.6$&$30.0$&$\mathbf{12.0}$&$32.2$&$43.3$&$74.7$ \\ \cline{2-9}
%    & ATSS w. maskIoU & $\mathbf{30.8}$& $\mathbf{50.6}$ & $\mathbf{32.2}$ & $12.5$ & $\mathbf{33.6}$ & $\mathbf{47.7}$ & $\mathbf{73.4}$ \\
    & ATSS w. maIoU (Ours)& $\mathbf{30.4}$& $\mathbf{50.3}$ & $\mathbf{31.4}$ & $11.5$ & $\mathbf{33.9}$ & $\mathbf{46.3}$ & $\mathbf{73.7}$ \\ \hline \hline 
    \multirow{5}{*}{700} &fixed IoU threshold&$29.7$&$50.0$&$30.4$&$14.2$&$32.8$&$43.7$&$74.3$\\
    & ATSS w. IoU&$30.8$&$51.8$&$31.2$&$14.1$&$35.0$&$44.0$&$73.3$\\
    &ATSS w. DIoU&$30.9$&$51.9$&$31.7$&$14.0$&$35.4$&$44.0$&$73.3$ \\
    &ATSS w. GIoU&$30.1$&$50.7$&$31.0$&$14.0$&$33.8$&$43.1$&$74.0$ \\ \cline{2-9}
    & ATSS w. maIoU (Ours)&$\mathbf{31.8}$&$\mathbf{52.8}$&$\mathbf{32.8}$&$\mathbf{14.7}$&$\mathbf{35.6}$&$\mathbf{45.7}$&$\mathbf{72.5}$\\ \hline 
    \end{tabular}
\end{table}
\label{subsec:Ablation}

\subsection{Ablation Experiments}
In this section, we demonstrate that our maIoU improves upon other assigners based on IoU variants consistently and our Alg. \ref{alg:MaskAwareIoU} makes computation of maIoU feasible during training.

\noindent \textbf{Using ATSS with IoU.} We first replace the fixed IoU threshold assigner of YOLACT by ATSS with (w.) IoU and have a stronger baseline for our maIoU: ATSS w. IoU improves fixed IoU assigner by $0.5 - 1.1$ mask AP over different scales (Table \ref{tab:minival}).

\noindent \textbf{Replacing IoU of ATSS with maIoU.} Replacing IoU in ATSS by our maIoU  (Table \ref{tab:minival}) (i) improves fixed IoU assigner by $1.3$, $1.9$ and $2.1$ mask APs for 400, 550 and 700 scales respectively, (ii) outperforms ATSS w. IoU  by $\sim 1.0$ mask AP in all scales, (iii) performs also better than other IoU variants (i.e. ATSS w. GIoU and ATSS w. DIoU). We note that the contribution of maIoU (i) on models trained by images with larger scales (700 vs. 400 in Table \ref{tab:minival}) and (ii) on larger objects ($\mathrm{AP_{L}}$ vs. $\mathrm{AP_{S}}$) are more significant. This is intuitive since the shape of the object gets more precise as the object gets larger. 

\noindent \textbf{Computing maIoU Efficiently.} Computing maIoU for every anchor-ground truth pair during training by brute force is infeasible, i.e. it would take $\sim 3$ months to train a single model with $41.89$ sec/iteration (Table \ref{tab:runtime}). Using Alg. \ref{alg:MaskAwareIoU}, we reduce the average iteration time by $\sim 70\times$ to $0.59$ sec/iteration, which is similar to other standard assigners (Table \ref{tab:runtime}).

\begin{table}
\RawFloats
\parbox{.37\linewidth}{
    \centering
    \small
    \setlength{\tabcolsep}{0.3em}
    \caption{Avg. iteration time ($t$) of assigners. While brute force maIoU computation is inefficient (Alg. \ref{alg:MaskAwareIoU} is \xmark); our Alg. \ref{alg:MaskAwareIoU} decreases $t$ by $\sim 70 \times$ and has similar $t$ with Fixed IoU Thr. and ATSS w. IoU.}
    \label{tab:runtime}
    \begin{tabular}{|c|c|c|} \hline
    Assigner& Alg. \ref{alg:MaskAwareIoU}& $t$ (sec.) \\ \hline
    Fixed IoU Thr.&N/A&$0.51$\\
    %0.5255, 0.5089, 
    ATSS w. IoU&N/A&$0.57$ 
    %(1.12x)
    \\
    %0.8141, 0.8030, 
    ATSS w. maIoU& \xmark &$41.89$ 
    %(82.14x)
    \\ \hline
    ATSS w. maIoU &\cmark &$0.59$ 
    %(1.16x)
    \\
    %$0.5065, 0.4888$
     \hline
    \end{tabular}
}
\hfill
\parbox{.61\linewidth}{
    \centering
    \small
    %\footnotesize
    \setlength{\tabcolsep}{0.4em}
    \caption{ATSS w. maIoU (underlined) makes YOLACT more accurate and  $ \sim 25 \%$ faster mainly owing to less anchors. Thanks to this efficiency, we build maYOLACT-550 with $34.8$ AP and still larger fps than YOLACT. 
    %Using images with 700 scale, we reach $37.2$ mask AP at $25$fps.
    }
    \label{tab:mayolact}
    \begin{tabular}{|c|c|c|c|c|c|} \hline
    \multicolumn{2}{|c|}{Method}&$\mathrm{AP}$&$\mathrm{AP^{box}}$& fps & Anchor \#  \\ \hline
    \multirow{6}{*}{\rotatebox{90}{\footnotesize{maYOLACT-550}}}&YOLACT-550&$28.5$&$30.7$&$27$&$\sim 19.2 K$ \\ \cline{2-6}
    &+ \underline{ATSS w. maIoU}&\underline{$30.4$}&\underline{$32.5$} &\underline{$\mathbf{33}$}&\underline{$\mathbf{\sim 6.4 K}$}   \\ 
    &+ Carafe FPN \cite{carafe} &$31.4$&$33.3$&$32$&$\mathbf{\sim 6.4 K}$  \\
    &+ DCNv2 \cite{DCNv2} & $33.2$ &$35.8$& $31$ &$\mathbf{\sim 6.4 K}$ \\
    &+ more anchors &$33.5$&$36.3$&$30$&$\sim 12.8 K$\\
    &+ cosine annealing \cite{sgdr}  & $34.8$&$37.9$& $30$ &$\sim 12.8 K$\\ \hline
%    \multicolumn{2}{|c|}{maYOLACT-700}&$\mathbf{37.2}$&$\mathbf{40.4}$&$25$ &$\sim 12.8 K$\\
%    \multirow{6}{*}{\rotatebox{90}{\footnotesize{maYOLACT-550}}}&YOLACT-550&$28.5$& &$26.6$ & $\sim 19.2 K$ \\ \cline{2-5}
%    &+ ATSS w. maIoU&$30.4$& &$\mathbf{33.1}$&$\mathbf{\sim 6.4 K}$   \\ 
%    &+ Carafe FPN \cite{carafe} &$31.4$& &$32.1$&$\mathbf{\sim 6.4 K}$  \\
%    &+ DCNv2 \cite{DCNv2} & $33.2$ & & $30.9$ &$\mathbf{\sim 6.4 K}$ \\
%    &+ more anchors &$33.5$& & $30.2$ & $\sim 12.8 K$\\
%    &+ cosine annealing \cite{sgdr}  & $34.8$& & $30.9$ &$\sim 12.8 K$\\ \hline
%    \multicolumn{2}{|c|}{maYOLACT-700} & $\mathbf{37.2}$& & $24.7$ &$\sim 12.8 K$\\
%    \hline
    \end{tabular}
}
\end{table}

\subsection{maYOLACT Detector: Faster and Stronger}
%This section shows that ATSS w. our maIoU also decreases the inference time of YOLACT and builds maYOLACT as a more accurate real-time detector, i.e. inference time $\geq 25$fps.

Thanks to using fewer number of anchors, YOLACT trained with our ATSS w. maIoU assigner (underlined in Table \ref{tab:mayolact}) is $\sim 25 \%$ faster than baseline YOLACT ($33$ vs. $27$ fps\footnote{For all models, we follow and report the results on the  mmdetection framework \cite{mmdetection} on a single Tesla V100 GPU. Mmdetection's YOLACT is slower than the official implementation by Bolya et al. \cite{yolact}, who reported $45$fps.}), pointing out the importance of anchor design for the efficiency of real-time systems as well\footnote{Note that more efficient models can be obtained by using better anchor design methods \cite{MetaAnchor,AdaptiveAnchor1,AdaptiveAnchor3,AdaptiveAnchor2}}. Exploiting this run-time gap; our aim in this section is to extend the standard YOLACT using our maIoU and the recent improvements in order to make it competitive with the recent methods also by keeping the resulting detector to process images in real-time\footnote{We use 25fps as the cut-off for ``real-time'' following the common video signal standards (e.g. PAL \cite{pal} and SECAM \cite{secam}) and existing methods \cite{bpd, salientod, basnet, tinieryolo}.}. To achieve that, we use (i) carafe \cite{carafe} as the upsampling operation of FPN \cite{FeaturePyramidNetwork}, (ii) deformable convolutions \cite{DCNv2} in the backbone, (ii) two anchors with base scales 4 and 8 on each pixel, and (iv) cosine annealing  with an initial learning rate of $0.008$ by replacing the step learning rate decay. Effect of these improvements are presented in Table \ref{tab:mayolact} and the resulting detector with these improvements is coined as \textit{maYOLACT}. Note that our maYOLACT-550 detector is still faster than baseline YOLACT-550 and improves it by $+6.3$ mask AP and $+7.2$ box AP reaching $34.8$ mask AP and $37.9$ box AP (Table \ref{tab:mayolact}). 
% With 700 scale images, maYOLACT yields $37.2$ mask AP and $40.4$ box AP at $25$ fps. 
%
\begin{table}
    \centering
    \small
    \setlength{\tabcolsep}{0.4em}
    \caption{Comparison with SOTA on COCO \textit{test-dev}. Our maYOLACT-700 establishes a new SOTA for real-time instance segmentation. 
    %For YOLACT-550+ATSS w. IoU baseline, 
    $^*$ implies our implementation for YOLACT with ATSS w.IoU.
    %and $^{**}$ includes improvements (Carafe FPN, DCNv2, more anchors and cosine annealing). 
    When a paper does not report a performance measure, N/A is assigned and we reproduce the performance using its repository for completeness (shown by $^\dagger$). }
    \label{tab:testdev}
    \begin{tabular}{|c|c|c|c|c|c|c|c|c|c|} \hline
        \multicolumn{2}{|c|}{Methods}&Backbone&$\mathrm{AP}$& $\mathrm{AP_{50}}$ & $\mathrm{AP_{75}}$ & $\mathrm{AP_{S}}$ & $\mathrm{AP_{M}}$ & $\mathrm{AP_{L}}$&Reference \\ \hline
    %\textit{Methods w. fps $ < 25$} & & & & & & &\\
    \multirow{5}{*}{\rotatebox{90}{fps $ < 25$}}
    &YOLACT-700 \cite{yolact} & ResNet-101&$31.2$&$50.6$&$32.8$&$12.1$&$33.3$&$47.1$&ICCV 19\\    
    &PolarMask \cite{polarmask}& ResNet-101&$32.1$&$53.7$&$33.1$&$14.7$&$33.8$&$45.3$&CVPR 20\\
    &PolarMask++ \cite{PolarMask-plus}& ResNet-101&$33.8$&$57.5$&$34.6$&$16.6$&$35.8$&$46.2$&TPAMI 21\\
    &RetinaMask \cite{retinamask}& ResNet-101&$34.7$&$55.4$&$36.9$&$14.3$&$36.7$&$50.5$&Preprint\\
    &Mask R-CNN \cite{tensormask}& ResNet-50&$36.8$&$59.2$&$39.3$&$17.1$&$38.7$&$52.1$&ICCV 17\\ 
    &TensorMask \cite{tensormask}& ResNet-101&$37.1$&$59.3$&$39.4$&$17.4$&$39.1$&$51.6$&ICCV 19\\ \hline \hline
    %\textit{Methods w. fps $ \geq 25$} & & & & & & &\\
    \multirow{8}{*}{\rotatebox{90}{fps $ \geq 25$}}&YOLACT-550 \cite{yolact} & ResNet-50&$28.2$&$46.6$&$29.2$&$9.2$&$29.3$&$44.8$&ICCV 19\\
    &YOLACT-550$^*$& ResNet-50&$29.7$&$49.9$&$30.7$&$11.9$&$32.4$&$42.7$&Baseline\\
    &Solov2-448 \cite{solov2} & ResNet-50 &$34.0$&$54.0$&$36.1$&N/A&N/A&N/A&NeurIPS 20\\
    &Solov2-448$^\dagger$ \cite{solov2-imp} & ResNet-50 &$34.0$&$54.0$&$36.0$&$10.3$&$36.3$&$53.8$&NeurIPS 20\\
    &YOLACT-550++ \cite{yolact-plus} & ResNet-50&$34.1$&$53.3$&$36.2$&$11.7$&$36.1$&$53.6$&TPAMI 20\\
    &YOLACT-550++ \cite{yolact-plus} & ResNet-101&$34.6$&$53.8$&$36.9$&$11.9$&$36.8$&$55.1$&TPAMI 20\\
%    &YOLACT-550$^{**}$& ResNet-50&$34.9$&$56.0$&$36.8$&$14.5$&$38.0$&$50.5$&Baseline\\
    % BlendMask-RT \cite{blendmask} & ResNet-50&$35.1$&$53.3$&$36.2$&$11.7$&$36.1$&$53.6$\\
    &CenterMask-Lite$^\dagger$ \cite{centermask-imp} & VoVNetV2-39 &$35.7$&$56.7$&$37.9$&$\mathbf{18.4}$&$37.8$&$47.3$&CVPR 20\\
    &CenterMask-Lite \cite{centermask} & VoVNetV2-39 &$36.3$&N/A&N/A&$15.6$&$38.1$&$53.1$&CVPR 20\\
    &Solov2-512$^\dagger$ \cite{solov2-imp} & ResNet-50 &$36.9$&$57.5$&$39.4$&$12.8$&$39.7$&$\mathbf{57.1}$&NeurIPS 20\\
    &Solov2-512 \cite{solov2} & ResNet-50 &$37.1$&$57.7$&$39.7$&N/A&N/A&N/A&NeurIPS 20\\
\cline{2-10}
    %\textit{Ours (fps $ \geq 25$)} & & & & & & &\\
    &maYOLACT-550 (Ours) & ResNet-50&$35.2$ &$56.2$&$37.1$& $14.7$&$38.0$&$51.4$ &\\
    &maYOLACT-700 (Ours) & ResNet-50&$\mathbf{37.7}$ &$\mathbf{59.4}$&$\mathbf{39.9}$& $18.1$&$\mathbf{40.8}$&$52.5$ &\\
    %maYOLACT-550 (Ours)&ResNet-101&$\mathbf{32.1}$&$\mathbf{52.8}$&$\mathbf{33.5}$& $\mathbf{12.9}$&$\mathbf{34.9}$&$\mathbf{47.5}$ \\ 
    \hline
    \end{tabular}
\end{table}
\subsection{Comparison with State-of-the-art (SOTA)}
Table \ref{tab:testdev} compares our maYOLACT with state-of-the-art methods on COCO \textit{test-dev}.

\noindent \textbf{Comparison with YOLACT variants.} Achieving $35.2$ mask AP, our maYOLACT-550 outperforms all YOLACT variants including the ones with larger backbones (e.g. YOLACT-550++ with ResNet-101) and larger scales (e.g. YOLACT-700). Besides, different from YOLACT++ \cite{yolact-plus}, which is $\sim 25\%$ slower than YOLACT (see Table 6 in Bolya et al. \cite{yolact-plus}), our maYOLACT-550 is faster than YOLACT-550 (Table \ref{tab:mayolact}), and still keep $7$ mask AP gain also on COCO \textit{test-dev} reaching $35.2$ mask AP (Table \ref{tab:testdev}).

\noindent \textbf{Comparison with real-time methods.} Without multi-scale training as in Solov2 \cite{solov2} or specially designed backbone as in CenterMask \cite{centermask}; our maYOLACT-700 reaches $37.7$ mask AP at 25fps and outperforms existing real-time counterparts. Besides, our best model achieves $59.4$ wrt. common $\mathrm{AP_{50}}$ metric with a gap of $1.7$ $\mathrm{AP_{50}}$ points compared to its closest real time counterpart (i.e. SOLOv2-512).  

\noindent \textbf{Comparison with other methods.} Our maYOLACT is also competitive against slower methods (Table \ref{tab:testdev}): It outperforms PolarMask++ \cite{PolarMask-plus}, RetinaMask \cite{retinamask}, Mask R-CNN \cite{MaskRCNN} and TensorMask \cite{tensormask} while being faster. To illustrate, on Tesla V100 GPU, our maYOLACT-700 (i) has $\sim 2\times$ more throughput with $25$fps and nearly $4$ mask AP gain ($37.7$ AP - Table \ref{tab:testdev}) compared to PolarMask++ on ResNet-101 with $14$ fps test time;  and (ii) is $\sim 8 \times$ faster than TensorMask on ResNet-101 (i.e. $\sim 3$fps) with similar performance.

%% file: sections/5.Conclusions.tex
\section{Conclusion}
\label{sec:Conclusion}
We presented maIoU to assign a proximity value for an anchor compared to both a ground truth box and its mask. Using maIoU to assign anchors as positive or negative for training instance segmentation methods, we utilised the shape  of  objects as provided by the ground-truth segmentation masks. We showed that ATSS with our maIoU also improves throughput of the model. Exploiting this efficiency, we improved the performance further and reached SOTA results in real-time.